%% file: main.tex
\documentclass[runningheads]{llncs}

\usepackage{eccv}

\input{preamble}

\usepackage{eccvabbrv}

\usepackage{graphicx}
\usepackage{booktabs}

\usepackage[accsupp]{axessibility}  

\usepackage{hyperref}

\usepackage{orcidlink}

\begin{document}

\title{FaCT-GS: Fast and Scalable CT Reconstruction with Gaussian Splatting}

\author{Pawel Tomasz Pieta \orcidlink{0009-0005-7634-6627} \and
Rasmus Juul Pedersen \orcidlink{0009-0007-8395-4270} \and
Sina Borgi \orcidlink{0000-0001-7404-7766} \and
Jakob Sauer Jørgensen \orcidlink{0000-0001-9114-754X} \and
Jens Wenzel Andreasen \orcidlink{0000-0002-3145-0229} \and
Vedrana Andersen Dahl \orcidlink{0000-0001-6734-5570} }

\authorrunning{P.T.~Pieta et al.}

\institute{Technical University of Denmark, 
Kgs. Lyngby, Denmark
\email{\{papi,rjupe,borgi,jakj,jewa,vand\}@dtu.dk}}

\maketitle

\begin{abstract}
  Gaussian Splatting (GS) has emerged as a dominating technique for image rendering and has quickly been adapted for the X-ray Computed Tomography (CT) reconstruction task. However, despite its growing popularity, the benefits of GS are typically not substantial enough to motivate a transition from well-established reconstruction algorithms. This paper addresses the most significant remaining limitations of the GS-based approach by introducing FaCT-GS, a framework for fast and flexible CT reconstruction. Enabled by an in-depth optimization of the voxelization and rasterization pipelines, our new method is significantly faster than its predecessors and scales well with projection and output volume size. Furthermore, the improved voxelization enables rapid fitting of Gaussians to pre-existing volumes, which can serve as a prior for warm-starting the reconstruction, or simply as an alternative, compressed representation. FaCT-GS is over 4$\times$ faster than the State of the Art GS CT reconstruction on standard $512^2$ projections, and over 13$\times$ faster on 2k projections. Implementation and data available through: \url{https://papieta.github.io/fact-gs/}.
\end{abstract}

\section{Introduction} \label{sec:intro}

In X-ray Computed Tomography (CT), the reconstructed 3D volume is obtained from a set of acquired X-ray projection images (or views), with reconstruction quality strongly dependent on the number of available projections. In sparse-view tomography, the number of projections is small, requiring specialized algorithms to achieve the best possible reconstruction quality.

\begin{figure}[!t]
\centering
\includegraphics[width=\columnwidth]{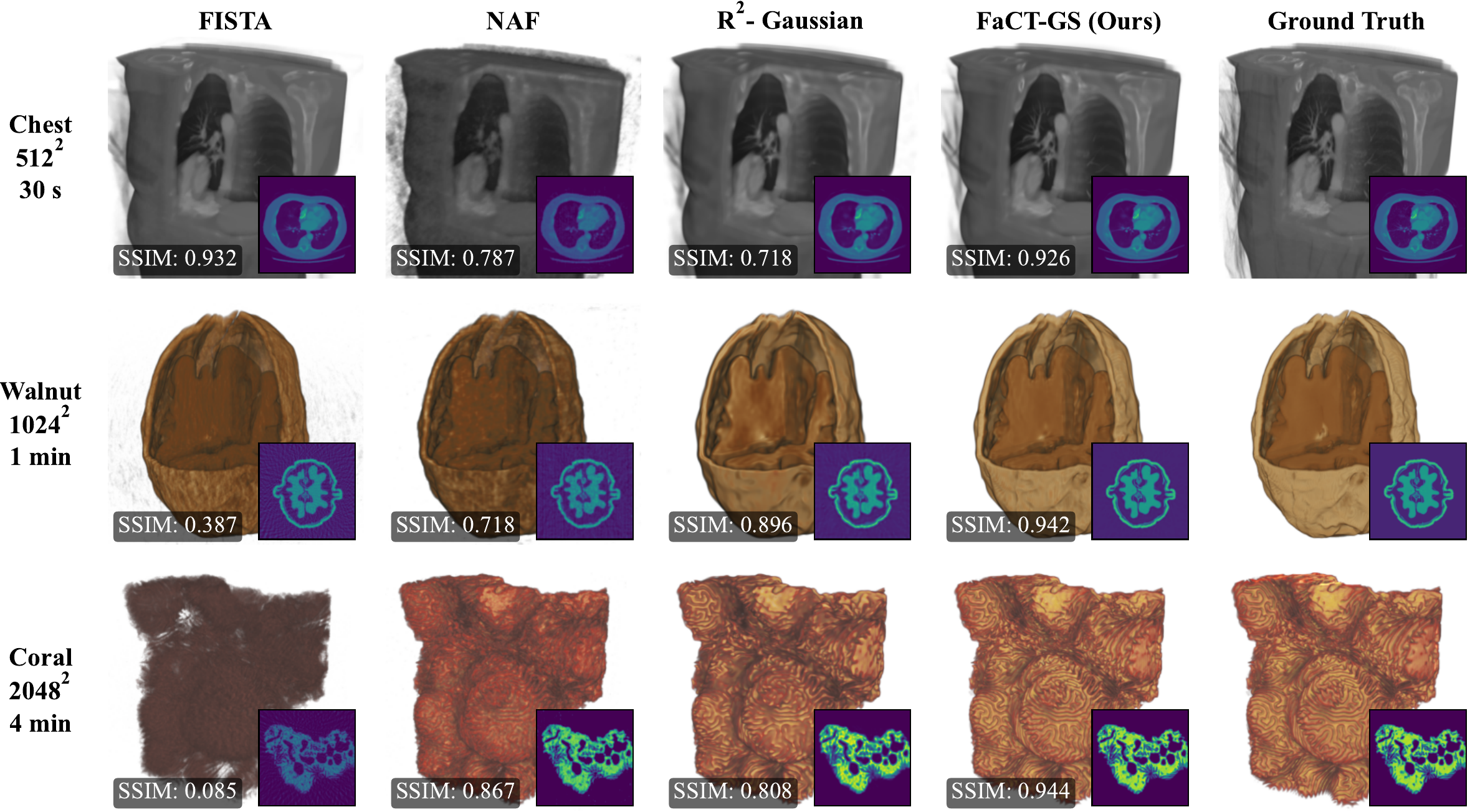}
\caption{Reconstruction results of select baselines and FaCT-GS given a fixed time for a certain problem size.
 FaCT-GS provides superior CT reconstruction capabilities and scales exceptionally well with increasing projection size.}
\label{fig:teaser_figure}
\end{figure}

Classical sparse-view reconstruction algorithms~\cite{sirt_gilbert1972a,sart_andersen1984a,asd_pocs_sidky2008a,fista_beck2009a} provide fast results, but often suffer from residual artifacts and reduced fidelity, motivating continuous research into alternative, typically more advanced, approaches. Recently, novel scene representation techniques have been adapted to CT data~\cite{intratomo_zang2021a,NAF_zha2022a,sax_nerf_cai2024a,neat_ruckert2022a,zha2024r} where they demonstrated a substantial improvement in reconstruction quality, albeit at a significantly increased computational cost. The current State of the Art (SotA) approach, R$^2$-Gaussian~\cite{zha2024r}, is one of the few papers to directly address the performance-quality tradeoff by implementing a complete Gaussian Splatting-based CT reconstruction pipeline, achieving both much faster execution time than its predecessors and better performance. Despite these advancements, learning-based approaches are not mature enough to serve as a practical alternative to classical methods. Even R$^2$-Gaussian requires more than $9\times$ longer execution than a well-established and effective FISTA~\cite{fista_beck2009a}, when reconstructing from a set of 75 projections of size 512$\times$512 pixels. While such resolution is standard in diagnostic imaging, modern scanners routinely operate at 1k and 2k resolutions, substantially increasing the computational burden of the reconstruction.

To facilitate the adoption of modern sparse-view reconstruction techniques, their performance must be better aligned with the practical requirements of the CT community. Motivated by this need, we introduce FaCT-GS, a fast and scalable Gaussian Splatting-based CT reconstruction pipeline. Inspired by recent advances in Gaussian Splatting acceleration, we re-implement and accelerate all core components of the reconstruction process, reducing execution time to approximately 2 minutes per scan (on 75 projections of $512^2$ size), with competitive results after only 45 seconds. Moreover, our pipeline exhibits substantially improved scalability, generating high-quality reconstructions in about 1 minute for 1k projections and 4 minutes for 2k projections (\cref{fig:teaser_figure}).

We also redesign the model initialization strategy and propose two novel schemes: one based on the gradient of an FDK reconstruction, and one leveraging a pre-existing volumetric prior. The latter brings us closer to the capabilities of most classical algorithms, enabling an efficient warm start of the optimization process. Finally, we highlight the compression properties of GS models, demonstrating that fitting them to a volume results in a quality–compression tradeoff comparable to representing the data as a stack of 2D \texttt{JPEG} images.

\section{Background and Related Work} \label{sec:related_work}

\textbf{CT imaging and reconstruction} \quad 
When capturing X-ray projection data, the X-ray beam passes the scanned object on its way from the X-ray source to the detector. The beam is attenuated by the constituents of the scanned object, and the resulting intensity measured at the detector reflects the total attenuation along the beam path. By collecting projections from many directions -- typically at least several hundred -- it is possible to reconstruct a dense 3D representation of the object. Although CT reconstruction is an inverse and ill-posed problem, under favorable conditions, a fast and deterministic Feldkamp–Davis–Kress (FDK) algorithm~\cite{feldkamp1984a}  can yield highly accurate volumetric reconstructions.

In many practical scenarios, the number of collected projections has to be limited. Each projection costs time and contributes to the cumulative dose of ionizing radiation -- a problem especially critical in medical imaging~\cite{brenner2007a,mccollough2009a}. Additionally, physical constraints of the scanning geometry may restrict the angular coverage or projection count~\cite{gao2007a,jin2010a,longo2022a}. When the number of projections is insufficient -- so-called sparse-view tomography -- FDK reconstructions suffer from pronounced artifacts and degraded image quality that are not directly caused by the projection data.

\textbf{Traditional sparse-view tomography} \quad
A wide range of algorithms has been proposed to address reconstruction under limited projection conditions. In the majority of cases, these methods rely on iterative optimization procedures that drive the solution toward a global optimum, leading to an inherent tradeoff between reconstruction quality and computational cost. Such iterative algorithms can be broadly divided into two groups: algebraic and learning-based. Classical algebraic methods, such as SIRT~\cite{sirt_gilbert1972a}/SART~\cite{sart_andersen1984a}, ASD-POCS~\cite{asd_pocs_sidky2008a}, and FISTA~\cite{fista_beck2009a}, formulate the tomographic inverse problem as a system of equations augmented with explicit smoothness or sparsity priors. Among these, FISTA is often preferred in practice due to its low computational cost (less than 1 minute per scan on $512^2$ px projections) and flexibility, allowing for non-differentiable regularization terms.

\textbf{Learning-based methods and NeRFs} \quad
With the rise of deep learning, the research focus shifted toward data-driven reconstruction approaches~\cite{siunet_dong2019a,anirudh2018a,learned_primaldual_adler2018a,coil_sun2021a,fbpconvnet_jin2017a}, often achieved through a combination of neural and standard reconstruction components. Importantly, they typically do not support a complete self-supervised framework -- a limitation which has been largely overcome through the recent advent of neural scene representations and Neural Radiance Fields~\cite{nerf_mildenhall2020a,henc_muller2022a,pixelnerf_yu2021a,dvgo_sun2022a,plenoctrees_yu2021a,msvnerf_chen2021a}, and their adaptation to CT~\cite{intratomo_zang2021a,NAF_zha2022a,pedersen2025a,maas2023a,AC_IND_xie2025a,mednerf_figueroa2022a,VolumeNeRF_liu2024a,neat_ruckert2022a,sax_nerf_cai2024a}. These methods represent the volume using compact MLP-based field models, enabling a simple and flexible abstraction of the reconstruction process. Here, most notable mentions are IntraTomo~\cite{intratomo_zang2021a}, NAF~\cite{NAF_zha2022a}, and NeAT~\cite{neat_ruckert2022a}, as the first successful implementations, and SAX-NeRF~\cite{sax_nerf_cai2024a}, as a slower, but best-performing model so far. Despite their impressive reconstruction quality, NeRF-based methods remain substantially slower than classical algebraic techniques, typically requiring 15–30 minutes per scan. Calculating each pixel in a CT projection requires multiple neural network queries along the path of the X-ray, limiting the scalability of these methods.

\textbf{Gaussian Splatting} \quad
Most recently, Gaussian Splatting (GS) methods have emerged as a powerful alternative for scene representation~\cite{kerbl2023a,zhang2025a,taming3dgs,keil2023a,tang2024a,guedon2024a,wu2024a,lu2024a,lee2024a}. The core of their approach is modelling a scene through a collection of optimized 3D Gaussians, offering several advantages over NeRFs. Most importantly, 3D Gaussians can be projected analytically onto a 2D plane, eliminating the expensive per-ray query process. To date, only a handful of works have explored Gaussian Splatting for CT imaging and reconstruction~\cite{zha2024r,yu2025a,gao2024a,li2025a,cai2025a,jecklin2025a,wang2025a}, with the most notable one being R$^2$-Gaussian~\cite{zha2024r} standing out as a complete pipeline tailored to the assumptions and requirements of CT imaging. Compared to NeRF-based methods,  R$^2$-Gaussian reports a reduced reconstruction time of approximately 10–15 minutes per scan, with satisfactory results available after just 3-4 minutes, positioning Gaussian Splatting as a particularly promising approach for CT reconstruction.

\textbf{Gaussian Splatting acceleration} \quad
A growing body of concurrent works focuses on accelerating the Gaussian Splatting pipeline~\cite{papantonakis2024a,soc2025a,girish2025a,fang2024a,fastergs_hahlbohm2026,taming3dgs,distwar_durvasula2023,wang2024a,stopthepop_radl2024a,speedysplat_hanson2025a}, primarily in the context of neural scene representation. While many of the proposed strategies are effective, only a subset can be directly transferred to CT reconstruction, often requiring careful adaptation and re-evaluation. In this work, we adopt several of the most established and impactful techniques, while also proposing improvements targeted specifically for CT reconstruction.

\section{Method} \label{sec:method}
Our overall CT reconstruction pipeline is built upon that of $R^2$-Gaussian~\cite{zha2024r}, which itself is derived from the original Gaussian Splatting for scene representation~\cite{kerbl2023a}. The attenuation field to be reconstructed is modelled with a collection of learnable 3D Gaussian kernels $G_i^3$, with each forming a density field:
\begin{equation}
    G_i^3\left(\mathbf{x} \mid \rho_i, \mathbf{p}_i, \boldsymbol{\Sigma}_i\right)=\rho_i  \exp \left(-\frac{1}{2}\left(\mathbf{x}-\mathbf{p}_i\right)^{\top} \boldsymbol{\Sigma}_i^{-1}\left(\mathbf{x}-\mathbf{p}_i\right)\right),
\end{equation}
where $i$ is the index of the Gaussian, $\mathbf{x} \in \mathbb{R}^3$ is a position in space, and $\rho_i$, $\mathbf{p}_i \in \mathbb{R}^3$ and $\boldsymbol{\Sigma}_i \in \mathbb{R}^{3\times3}$ are the learnable parameters representing density, mean and covariance of the Gaussian. In contrast to neural scene representation, X-ray attenuation is not view-dependent, so the represented attenuation at $\mathbf{x}$ can be directly computed through density summation of contributing splats:
\begin{equation}
    \sigma(\mathbf{x})=\sum_{i=1}^M G_i^3\left(\mathbf{x} \mid \rho_i, \mathbf{p}_i, \boldsymbol{\Sigma}_i\right) .
\end{equation}

\begin{figure}[!t]
\centering
\includegraphics[width=0.95\columnwidth]{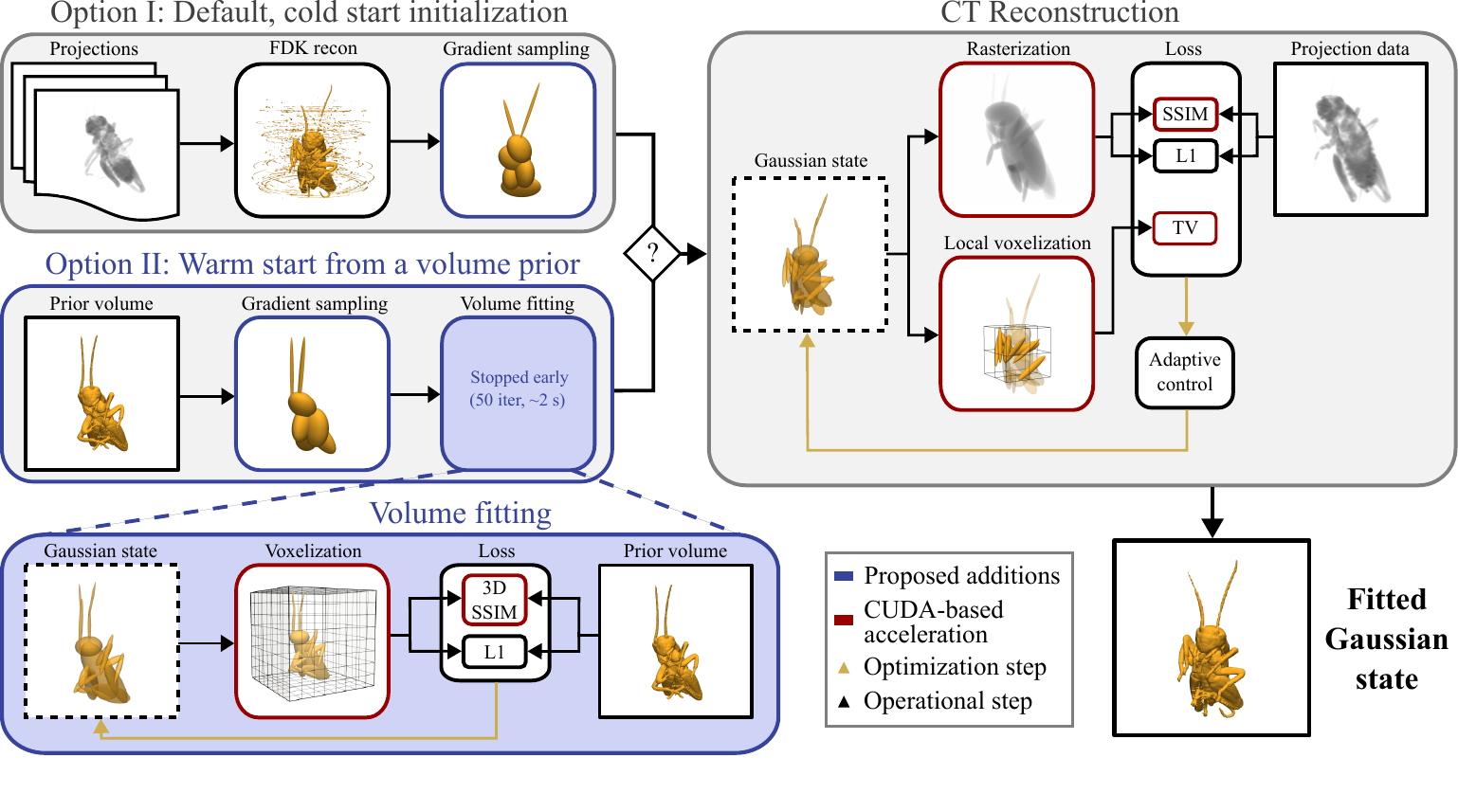}
\caption{
 FaCT-GS optimization pipeline. We propose two initialization schemes: one, based on the gradient of the FDK reconstruction from collected projections, and one based on the Gaussians fitted directly to a prior volume. We accelerate the core components of the pipeline: rasterization, voxelization, and compute-intensive loss functions.}
\label{fig:FaCT_GS_pipeline}
\end{figure}

\subsection{Initialization} \label{sec:method_initialization}

When no prior information is available, Gaussians are initialized from a low-quality FDK reconstruction, with background regions removed via thresholding. In existing approaches, Gaussian means are typically sampled with probability proportional to the reconstruction intensity~\cite{zha2024r,yu2025a}. We instead sample with probability proportional to the gradient magnitude, which ensures Gaussians are primarily located near high-frequency features~\cite{zhang2025a}. Gaussian densities are set based on the reconstructed intensities at the sampled positions and scaled down by a factor $k=0.15$. Each Gaussian is initialized with zero rotation (diagonal covariance matrix), and its scale is set equal to the nearest-neighbor distance.

\subsection{X-ray rasterization}

X-ray rasterization maps the set of Gaussians onto a discrete 2D image representing an X-ray projection. Each pixel value of a projection image is an integral of X-ray attenuation along a ray path $\mathbf{r}(t)$~\cite{natterer2001a}, which can be translated to a sum of integrals over the Gaussians intersected by the ray:
\begin{equation}\label{eq:xray_rasterization_1}
    I(\mathbf{r})=\sum_{i=1}^M \int G_i^3\left(\mathbf{r}(t) \mid \rho_i, \tilde{\mathbf{p}_i},\tilde{\boldsymbol{\Sigma}_i}\right) d t , \quad \mathbf{r}(t)=\tilde{\mathbf{x}}_0 + t\tilde{\mathbf{d}},
\end{equation}
where $I(\mathbf{r})$ denotes the rendered pixel value, and $\tilde{\mathbf{x}}$, $\tilde{\mathbf{p}_i}$, $\tilde{\boldsymbol{\Sigma}_i}$ and $\tilde{\mathbf{d}}$ are the $\mathbf{x},\mathbf{p}_i$, $\boldsymbol{\Sigma_i}$ and ray direction projected into the ray/detector space. CT acquisition is commonly described using either a parallel-beam geometry, where rays are perpendicular to the detector plane, or a cone-beam geometry, where rays emanate from a single small source. The former involves simple view transformations, while the latter is typically described with a pinhole camera model~\cite{natterer2001a,feldkamp1984a}.

Since a projection of a 3D Gaussian to a plane yields a 2D Gaussian, \cref{eq:xray_rasterization_1} can be simplified to:
\begin{equation}
    I(\hat{\mathbf{x}})=\sum_{i=1}^M G_i^2\left(\hat{\mathbf{x}} \left\lvert\, \mu_i \rho_i\right., \hat{\mathbf{p}}_i, \hat{\boldsymbol{\Sigma}}_i\right), 
\end{equation}
 where $\hat{\mathbf{x}}$, $\hat{\mathbf{p}}_i$ and $\hat{\boldsymbol{\Sigma}}_i$ are the $\tilde{\mathbf{x}},\tilde{\mathbf{p}_i}$ and $\tilde{\boldsymbol{\Sigma_i}}$ projected onto the 2D detector plane, and $\mu_i$ is an integration factor introduced in~\cite{zha2024r}. For efficiency, contributions are accumulated only for Gaussians satisfying $|\hat{\mathbf{x}}-\hat{\mathbf{p}_i}| < 3\lambda_\text{MAX}(\hat{\boldsymbol{\Sigma}}_i)$, with $\lambda_\text{MAX}$ signifying the largest eigenvalue.

\subsection{Voxelization} \label{sec:method_voxelization}

Voxelization discretizes the 3D space into a regular grid and accumulates Gaussian contributions within each voxel, producing a dense volumetric representation of the reconstructed object. Algorithmically, voxelization is simpler than rasterization, as it avoids mapping and projecting the Gaussians onto a 2D plane. In practice, however, it is substantially more computationally expensive due to the cubic growth in the number of grid elements.

\subsection{Optimization Loop}

During reconstruction, the Gaussians are rasterized into a 2D image ($\mathbf{I}_\mathrm{r}$) and compared to the available projection data ($\mathbf{I}_\mathrm{gt}$) using L1 loss ($\mathcal{L}_1$) and SSIM loss ($\mathcal{L}_\text{SSIM}$)~\cite{wang2004a}. Similarly to~\cite{zha2024r}, at each iteration, we also voxelize a small, randomly sampled section of the volume ($\mathbf{V}_\mathrm{{px}^3}$), but increase the sampling size from $8^3$ to $32^3$. On that volume, we calculate 3D total variation ($\mathcal{L}_\text{TV}$) as a homogeneity regularization prior. The total loss at each step is thus:
\begin{equation}
\mathcal{L}_{\text {total}}=\mathcal{L}_1\left(\mathbf{I}_\mathrm{r}, \mathbf{I}_\mathrm{gt}\right)+\alpha_{\text {SSIM }} \mathcal{L}_{\text {SSIM }}\left(\mathbf{I}_\mathrm{r}, \mathbf{I}_\mathrm{gt}\right)+\alpha_\mathrm{TV} \mathcal{L}_\mathrm{TV}\left(\mathbf{V}_{32^3}\right),
\end{equation}
where $\alpha_\text{SSIM}$ and $\alpha_\text{TV}$ are SSIM and TV loss weigths, respectively. In the backward pass, the loss is used to prune, clone, or split the Gaussians (Adaptive Control), and to adjust their parameters based on the propagated gradient. A visualization of the reconstruction optimization loop can be found on~\cref{fig:FaCT_GS_pipeline}.

\newpage
\subsection{Code acceleration} \label{sec:code_acceleration}

\textbf{Voxelization and rasterization} \quad Prior works identify two primary performance bottlenecks in the original rasterization kernels: inefficient splat bounding and backward pass stalling~\cite{fastergs_hahlbohm2026,taming3dgs,distwar_durvasula2023,wang2024a,stopthepop_radl2024a,speedysplat_hanson2025a}. The former arises from the use of square bounding boxes for Gaussian-shaped ellipses, which substantially overestimates the affected region for anisotropic and transparent Gaussians. Here, we follow~\cite{stopthepop_radl2024a,speedysplat_hanson2025a,wang2024a} and constrain the range using axis-aligned, rectangular bounding boxes with factored-in density. The second bottleneck originates from the per-pixel backward pass calculation: neighboring pixels often contribute gradients to the same Gaussian, leading to frequent write conflicts and warp stalls. In line with~\cite{taming3dgs}, we address this issue by reformulating the backward pass in a per-Gaussian manner, thereby eliminating the majority of conflicts.

We further clean up and streamline the pipeline to better match the requirements of CT imaging. In standard GS, splat contributions are accumulated in a specific order to model view-dependent occlusion. In CT, however, the integration order along the ray has no effect on the measured signal. While this realization was already highlighted in R$^2$-Gaussian, the ordering logic was not actually removed from the pipeline, incurring unnecessary overhead. We therefore remove it entirely. Finally, we reduce memory traffic by coalescing function parameters into compact \texttt{float4} structs, combining position and radius in the voxelizer, and conics (projected covariances) with $\mu$ in the rasterizer.

\begin{figure*}[!t]
\centering
     \begin{subfigure}[b]{0.54\columnwidth}
         \includegraphics[width=\columnwidth]{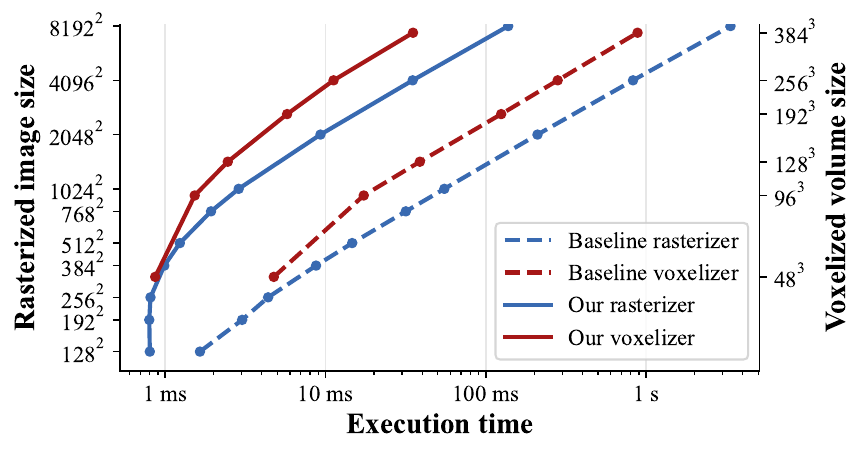}
         \caption{$N_\mathrm{G}=20000$, varying $N_\mathrm{p}$. }
         \label{fig:rast_vox_pix}
     \end{subfigure}
     \begin{subfigure}[b]{0.45\columnwidth}
         \includegraphics[width=\columnwidth]{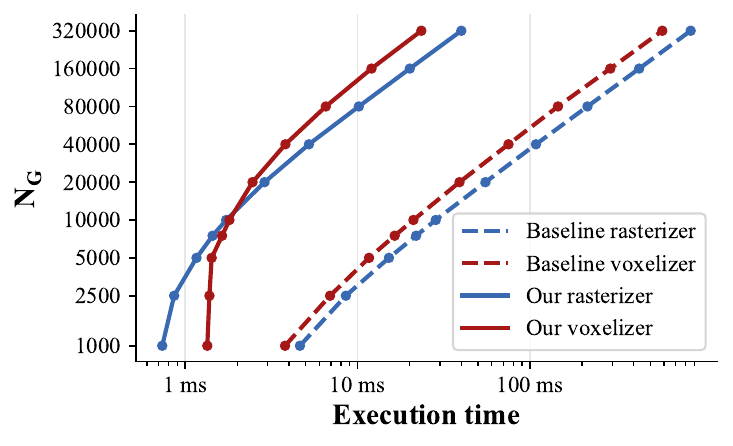}
         \caption{$N_\mathrm{px}\approx 1\times10^6$ , varying $N_\mathrm{G}$.}
         \label{fig:rast_vox_gauss}
     \end{subfigure}
\caption{Effective acceleration of the rasterization and voxelization components enabled through CUDA kernel optimization. Outlined relationships follow a power law. Plotted in log scale, baselines sourced from $R^2$-Gaussian~\cite{zha2024r}.}
\label{fig:rast_vox_optim}
\end{figure*}

\textbf{Kernel evaluation} \quad We evaluate the performance impact of the optimized kernels using a simple, synthetic test case, and compare them against the baseline implementations from $R^2$-Gaussian~\cite{zha2024r}. Gaussians are initialized with zero rotation and a uniform isotropic scale, and two test regimes are considered: (1) a fixed number of Gaussians ($N_\mathrm{G} = 20{,}000$) with varying image and volume resolutions (\cref{fig:rast_vox_pix}), and (2) a fixed image or volume size (approximately $N_\mathrm{px}\approx10^6$ pixels) with varying $N_\mathrm{G}$ (\cref{fig:rast_vox_gauss}).

Across all evaluated settings, execution times approximately follow power-law scaling, with the proposed kernels significantly outperforming the baseline rasterizer and voxelizer. For instance, on the same GPU (RTX 4070 Ti SUPER), a baseline rasterizer requires 210 ms to process a 2K image, while ours completes the same task in 9.3 ms, corresponding to a $22\times$ acceleration. The computational cost of voxelization discussed in~\cref{sec:method_voxelization} implies that, in typical scenarios, a voxelized volume contains up to $\sqrt{N_\mathrm{px}}$ times more elements than a rasterized projection. This relationship, combined with the observed power-law scaling, indicates that the proposed acceleration has the most profound impact on the voxelization stage.

\textbf{Optimization pipeline} \quad The accelerated GS pipeline is no longer dominated by the rasterization/voxelization step. A substantial fraction of the runtime is spent in the forward and backward propagation through the individual operations of the loss functions. By fusing these operations within dedicated CUDA kernels, it is possible to reduce memory traffic and launch overhead. We apply this strategy to the most computationally expensive loss components:
\begin{enumerate}
    \item We replace the 2D SSIM loss ($\mathcal{L}_\text{SSIM}$) calculation with the \texttt{fused-ssim}~\cite{taming3dgs} implementation, accelerating this step by $5-8$ times,
    \item We introduce a fused CUDA kernel for 3D SSIM loss, reaching an $11\times$ acceleration. Although inspired by \texttt{fused-ssim}~\cite{taming3dgs}, directly extending its logic to 3D would force an excessive use of thread block shared memory.  Instead, the depth dimension is processed using a ring buffer, with each thread operating on a single Z-axis row.
    \item We implement a custom fused CUDA kernel for the 3D total variation loss ($\mathcal{L}_\text{TV}$), reaching a $3\times$ acceleration for the $\mathbf{V}_{32^3}$ volume.
\end{enumerate}

\subsection{Volume prior initialization}

In established algebraic reconstruction methods such as SIRT or FISTA, the optimization can be warm-started from a volumetric prior (e.g., an existing reconstruction of the same or similar object), which -- depending on the degree of similarity -- can substantially accelerate convergence. For GS-based reconstruction, the most equivalent approach would either be to introduce a voxelizer-based Gaussian fitting stage (\cref{fig:FaCT_GS_pipeline}, Option II) or, more indirectly, to generate synthetic projections from the prior volume and pre-fit the GS model to this data. Both approaches have so far remained largely unexplored.

Following the optimization of the GS pipeline, voxelization becomes fast enough for regular use during fitting. This makes the voxelizer-based strategy both more direct and likely more effective. The procedure begins with a variant of the gradient-based Gaussian initialization (\cref{sec:method_initialization}), where gradients are extracted from the ground-truth volume ($\mathbf{V}_\mathrm{gt}$), followed by an optimization loop. At each step the current Gaussian representation is voxelized ($\mathbf{V}_\mathrm{r}$) and evaluated against $\mathbf{V}_\mathrm{gt}$ using $\mathcal{L}_1$ and a 3D variant of $\mathcal{L}_{\text {SSIM}}$ (weighted with $\alpha_{\text {SSIM}})$:
\begin{equation}
\mathcal{L}_{\text {total}}=\mathcal{L}_1\left(\mathbf{V}_\mathrm{r}, \mathbf{V}_\mathrm{gt}\right)+\alpha_{\text {SSIM }} \mathcal{L}_{\text {SSIM}}\left(\mathbf{V}_\mathrm{r}, \mathbf{V}_\mathrm{gt}\right).
\end{equation}

While -- due to the number of processed voxels -- a single step of this pipeline is slower than that of the reconstruction, the directness of the fitting process significantly decreases the number of necessary iterations. Furthermore, especially in the context of a reconstruction task, it may be neither necessary nor desirable to fully maximize the prior representation accuracy.
 
\section{Experiments} \label{sec:experiments}

\subsection{Performance evaluation} \label{sec:result_basic}

For evaluation, we compare FaCT-GS against a representative set of sparse-view CT reconstruction methods. From the traditional approaches, we choose FDK\cite{feldkamp1984a} as a baseline, SIRT~\cite{sirt_gilbert1972a} with a non-negativity constraint as a classical algebraic method, and FISTA~\cite{fista_beck2009a} with total variation regularization due to its effectiveness and widespread adoption in the CT community. From the learning-based approaches, we restrict our experiments to methods that provide a complete self-supervised pipeline. IntraTomo~\cite{intratomo_zang2021a} and NAF~\cite{NAF_zha2022a} are chosen to represent NeRF-based solutions, while R$^2$-Gaussian~\cite{zha2024r} serves as the current SotA Gaussian Splatting–based method. SIRT and FISTA are executed for 600 and 100 iterations, respectively, where one iteration corresponds to processing the full set of projections. Hyperparameters for the learning-based methods are kept consistent with their original implementations. For direct comparability, R$^2$-Gaussian and FaCT-GS are both initialized with 50,000 Gaussians and trained for 400 iterations. A small section of experiments was also run for SAX-NeRF~\cite{sax_nerf_cai2024a}. Although it outperformed IntraTomo and NAF in reconstruction quality, its execution time (3–4 hours per scan) renders it impractical for routine use. It is therefore excluded from further evaluation.
 
We perform reconstruction on the test dataset introduced in~\cite{zha2024r}, which comprises 15 real CT volumes with synthetically generated projections derived from publicly available datasets~\cite{roth2016pancreasct,x_plant,armato2011a,scivis}, as well as 3 cases with real projection data from the FIPS dataset~\cite{fips}. In all experiments, reconstructed volumes have a resolution of $256^3$. Projection sizes are set to $512^2$ for synthetic cases and $560^2$ for real acquisitions. Following~\cite{zha2024r}, we evaluate three sparse-view settings with 25, 50, and 75 projections. All experiments were run on the same machine with an NVIDIA H100 GPU using CUDA 12.8.1. The three traditional methods, FDK, SIRT, and FISTA, were run using the Core Imaging Library version 25\cite{CILv25,jorgensen2020} with TIGRE\cite{Biguri_2025} as the backend. 

\begin{table*}[t!]
\caption{Quantitative results on the $512^2$-projection dataset from \cite{zha2024r}. Colors signify \nI{best}, \nII{second-best}, and \nIII{third-best} numbers. FDK reconstruction was instantaneous. Qualitative comparison of the reconstructions can be found in \emph{Supplementary Material}.}
\centering
\begin{tabular}{@{}l@{\hspace{5pt}}ccc@{\hspace{10pt}}ccc@{\hspace{10pt}}ccc}
\toprule
   & \multicolumn{3}{c@{\hskip 10pt}}{75-view}  & \multicolumn{3}{c@{\hskip 10pt}}{50-view}  & \multicolumn{3}{c}{25-view} \\  \cmidrule(l{-1pt}r{10pt}){2-4}\cmidrule(l{-1pt}r{10pt}){5-7}\cmidrule(l{-1pt}r{-0.5pt}){8-10}
& PSNR & SSIM & Time  & PSNR & SSIM & Time & PSNR & SSIM & Time  \\ \midrule 
FDK  
& 26.76 & 0.388 & -  & 24.42 & 0.296 & - & 20.69 & 0.181 & - \\
SIRT  
& 33.96 & 0.844 & 2m13s  & 32.54 & 0.815 & 1m37s & 30.11 & 0.764 & 1m7s \\
FISTA  
& 37.07 & 0.934 & \nIIc 55s  & 35.85 & 0.923 & \nIIc 46s & 33.51 & 0.898 & \nIIIc 39s \\ \noalign{\vskip 0.5ex} \hdashline \noalign{\vskip 0.5ex}
IntraTomo  
& 29.67 & 0.829 & 29m24s  & 29.24 & 0.821 & 19m40s & 28.71 & 0.808 & 9m51s \\
NAF  
& 35.76 & 0.903 & 33m21s  & 34.41 & 0.891 & 22m14s & 31.61 & 0.849 & 11m8s \\
R$^2$-Gaussian 
& \nIIc 39.04 & \nIIc 0.946 & 9m  & \nIc 38.09 & \nIIc 0.939 & 5m46s & \nIIc 35.23 & \nIIIc 0.912 & 2m44s \\ \midrule
FaCT-GS (Ours)
& \nIc 39.05 & \nIc 0.950 & \nIIIc 1m52s  & \nIIc 37.97 & \nIc 0.943 & \nIIIc 1m15s & \nIc 35.33 & \nIc 0.92 & \nIIc 38s \\
FaCT-GS at 150 iter.
& \nIIIc 38.33 & \nIIc 0.946 & \nIc 43s  & \nIIIc 37.34 &  \nIIc 0.939 & \nIc 29s & \nIIIc 34.86 &  \nIIc 0.914 & \nIc 15s \\ \bottomrule
\end{tabular}
\label{tab:performance_basic}
\end{table*}

Quantitative performance results are summarized in~\cref{tab:performance_basic}. A full FaCT-GS optimization run requires at most 1 minute and 52 seconds (and only 38 seconds in the 25-view setting), achieving the highest SSIM and the second-highest PSNR across all methods. Compared to R$^2$-Gaussian, under the same number of iterations, FaCT-GS provides more than a $4.5\times$ speedup. Notably, after only 150 iterations, FaCT-GS already surpasses all competing methods in SSIM (matching R$^2$-Gaussian at 400 iterations), while requiring less than 45 s of fitting. This makes FaCT-GS the only learning-based approach that outperforms FISTA both in terms of speed and reconstruction quality.

\newpage
\subsection{Scaling capability} \label{sec:scaling_capability}

We choose the three best-performing methods from~\cref{tab:performance_basic} (FISTA, R$^2$-Gaussian, and  FaCT-GS) and evaluate how well their runtime scales with increasing projection and volume size. For this purpose, we use a large CT scan of a coral (\cref{fig:teaser_figure}), and downscale it to volume sizes ranging from $256^3$ to $1024^3$. From each volume instance, we generate 50 projections, each with the width of twice the volume width (from $512^2$ to $2048^2$) -- similarly to the data used in~\cref{sec:result_basic}.

As larger volumes contain proportionally more structural detail, both \linebreak FaCT‑GS and R$^2$‑Gaussian require an increased number of Gaussians to maintain reconstruction fidelity. We therefore double the Gaussian count for each doubling of projection/volume width, starting with 50,000 Gaussians for the projection size of $512^2$. For each reconstruction, we evaluate two scenarios:
\begin{enumerate}
    \item Training each method for the same, predefined number of 100 iterations ($N_\mathrm{iter}=100$) -- simulating a user manually comparing the methods.
    \item Training each method until it reaches 0.9 SSIM score -- testing the actual convergence capability of each method.
\end{enumerate}

\begin{figure*}[!t]
\centering
     \begin{subfigure}[b]{0.49\columnwidth}
         \includegraphics[width=\columnwidth]{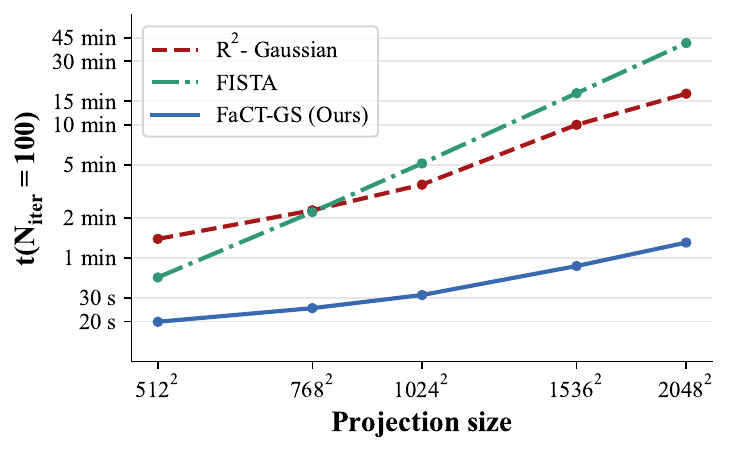}
         \caption{Time to reach 100 iterations. }
         \label{fig:time_to_iter}
     \end{subfigure}
     \begin{subfigure}[b]{0.49\columnwidth}
         \includegraphics[width=\columnwidth]{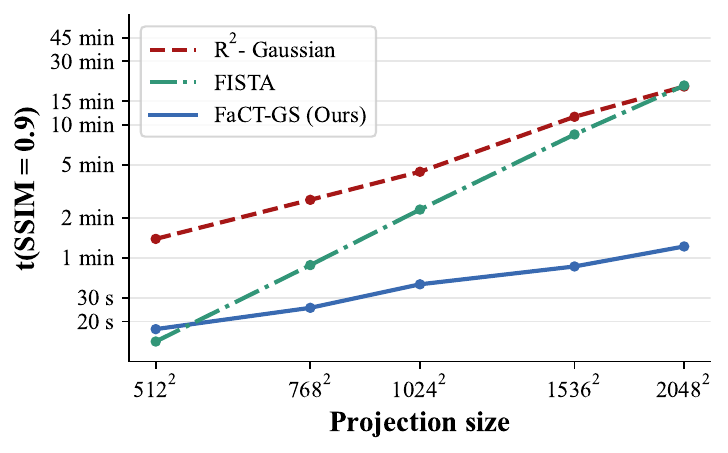}
         \caption{Time to achieve 0.9 SSIM.}
         \label{fig:time_to_ssim}
     \end{subfigure}
\caption{Impact of increasing volume and projection size on the performance of chosen methods. For larger data sizes, FaCT-GS performs exceptionally well, providing faster and better reconstruction than both FISTA and R$^2$-Gaussian at a fraction of the execution time. R$^2$-Gaussian iteration speed scales better with data size, but FISTA reaches convergence quicker.}
\label{fig:time_to}
\end{figure*}

In Scenario 1 (Fig.~\ref{fig:time_to_iter}), FaCT‑GS achieves the fastest iteration speed across all reconstruction sizes. At the smallest volume, it is already $2\times$ faster than FISTA and $4\times$ faster than R$^2$-Gaussian, and this advantage grows with increasing resolution. At the largest volume, FaCT‑GS provides a $31.5\times$ speed‑up over FISTA and a $13\times$ speed‑up over R$^2$-Gaussian, indicating that both competing methods scale substantially worse in iteration speed.
In Scenario 2 (Fig.~\ref{fig:time_to_ssim}), FISTA reaches an SSIM of 0.9 in the fewest iterations, which provides a modest advantage at the smallest reconstruction size. As projection and volume dimensions increase, this advantage is offset by FISTA’s steeply rising per‑iteration cost, causing its total runtime to grow faster than both R$^2$‑Gaussian and \linebreak FaCT‑GS. R$^2$‑Gaussian requires the most iterations, but its iteration time scales more moderately and its overall runtime intersects with FISTA at the largest tested volume. FaCT‑GS requires more iterations than FISTA but fewer than R$^2$‑Gaussian, while maintaining the lowest per‑iteration cost across all resolutions. Consequently, FaCT‑GS is the fastest method for all but the smallest reconstruction size, improving from $0.8\times$ to approximately $16\times$ FISTA’s speed as reconstruction size increases.

\subsection{Initialization evaluation} \label{sec:init_eval}

To assess the effectiveness of the two proposed initialization strategies, we construct a small dataset containing four realistic instances of volume pairs, where one volume serves as the reconstruction target and the other as a prior. The dataset includes two brain MRI scans and two pancreas CT scans from the Medical Segmentation Decathlon~\cite{med_dec_simpson2019}, two walnut CT scans—one from the R$^2$-Gaussian/FIPS dataset~\cite{zha2024r,fips} and one from HDTomo~\cite{jorgensen2021hdtomo} (also used in~\cref{fig:teaser_figure}), and a pair of cricket CT scans from the BugNIST dataset~\cite{jensen2024a}. All volumes are preprocessed using the same settings as in~\cref{sec:result_basic}, with 50 synthetic projections generated from each target volume. 

For each instance, we compute the 3D SSIM score between the target volume and the output from four initialization strategies: 
\begin{enumerate}
    \item \textbf{FDK Intensity}: Baseline intensity-based Gaussian mean sampling from the FDK reconstruction of the 50 projections.
    \item \textbf{FDK Gradient}: Our gradient-based sampling variant of the above.
    \item \textbf{Prior applied directly}: Direct use of the prior volume for initialization, as carried out in algebraic methods such as SIRT or FISTA.
    \item \textbf{FaCT-GS Rapid volume fit}: Our prior-based initialization via a dedicated rapid volume fitting stage.
\end{enumerate}
For the last method, the volume fitting pipeline is executed for 50 iterations, incurring an additional overhead of approximately 2 seconds.

\begin{figure}[!t]
\centering
\includegraphics[width=0.95\columnwidth]{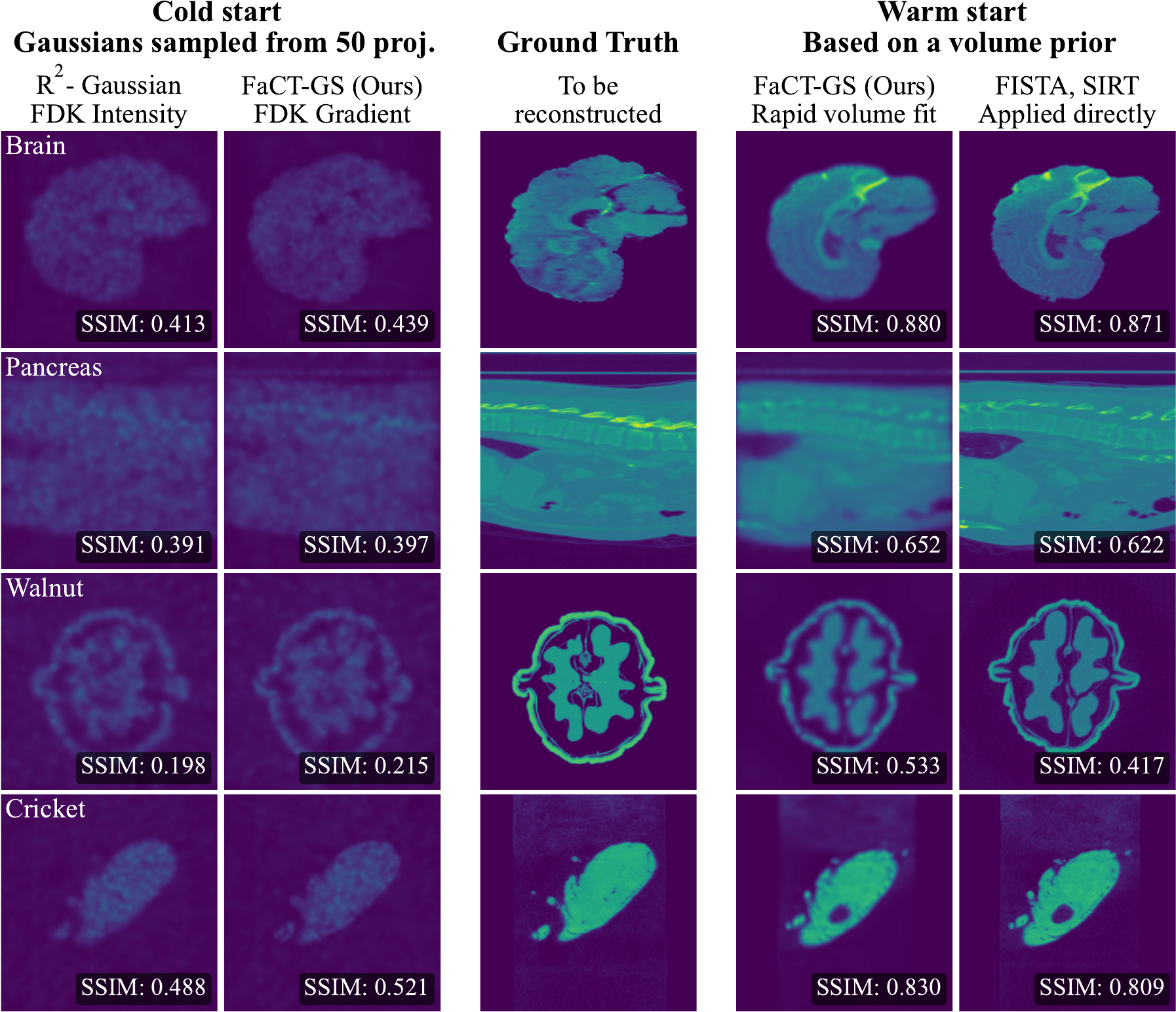}
\caption{Comparison of the output of reconstruction initialization strategies on four examples with a similar prior volume available. Gradient-based sampling better outlines the high-frequency features than intensity sampling. FaCT-GS rapid volume fitting demonstrates a similar impact as using the prior directly -- a standard approach in algebraic methods.}
\label{fig:init_comp}
\end{figure}

The results are shown in~\cref{fig:init_comp}. In all cases, the gradient-based approach provides an initial state with a moderately higher SSIM than intensity-based sampling, producing cleaner and more consistent delineation of edges and high-frequency structures. Both warm start strategies demonstrate the substantial benefit of incorporating a volumetric prior, achieving significantly higher SSIM scores than the FDK-based initializations. Our prior-integration strategy within the Gaussian Splatting pipeline attains the highest SSIM, indicating that the proposed approach successfully recreates the warm-start capability known from algebraic methods, and confirming that a rapid 50-iteration fitting stage is sufficient for effective reconstruction initialization.

\section{Volume fitting and compression} \label{sec:fit_compress}

Beyond enabling prior-based initialization, the volume fitting stage can be applied as a standalone optimization procedure, yielding a complete Gaussian representation of an arbitrary input volume. To demonstrate this capability, we perform volume fitting on all ground-truth volumes from the R$^2$-Gaussian dataset. Each model is optimized for 500 iterations using 40,000 Gaussians, resulting in an average fitting time of 19 seconds. 

The Gaussian representation can also serve as an efficient storage format. Each 3D Gaussian is parameterized by 11 values: a 3D position, 3D scale, 4D quaternion, and 1D density. Zhang et al.~\cite{zhang2025a} demonstrated in the context of 2D image compression that this representation can be quantized to \texttt{float16} with negligible quality degradation. Following this strategy, we reduce the storage size of each 3D splat to $11\times2=22$ bytes.

Importantly, in CT imaging, preserving information fidelity is often prioritized over storage efficiency, making volumetric compression a relatively underexplored area. Typically, volumes are stored either in raw \texttt{float32} format, converted to \texttt{uint8} (optionally combined with ZIP compression), or serialized as stacks of 2D \texttt{JPEG} images. More advanced and specialized compression techniques have been proposed~\cite{ihm1999a,lu2021a,hosny2020a,nguyen2001a}, but they are rarely used in practice. 

For GS-based CT reconstruction, however, the Gaussian representation already is the most information-rich form of the reconstructed data. Given the proposed fast voxelizer, directly storing the Gaussian representation and voxelizing it on demand becomes a natural and efficient strategy. We therefore use the volume fitting task as a proxy for evaluating the compression properties of the Gaussian representation. We compare its storage size and representation quality against the raw volume as well as two practical baselines: ZIP-compressed \texttt{uint8} volumes and stacks of 2D \texttt{JPEG} images.

\begin{figure}[t]
\centering
\begin{minipage}{\textwidth}
   \begin{minipage}[c]{0.48\textwidth}
    \centering
    \includegraphics[width=0.99\columnwidth]{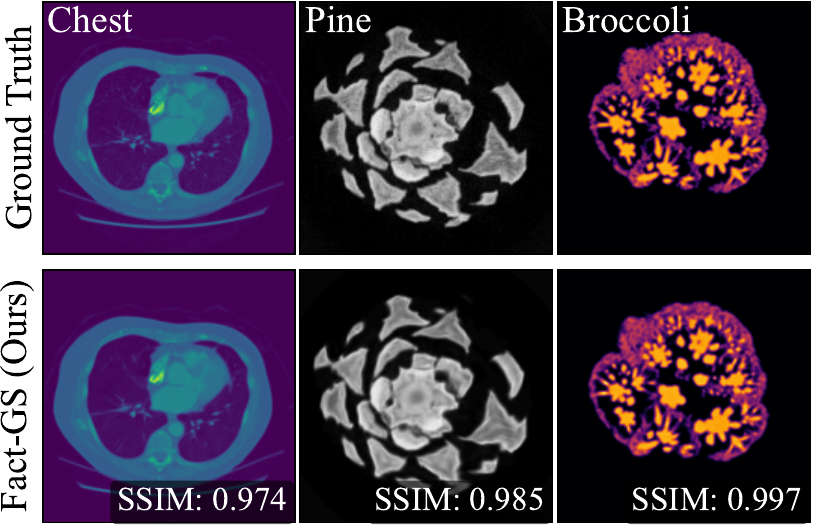}
    \captionof{figure}{Qualitative examples of the volume fitting result.}
    \label{fig:compression}
    \end{minipage}
\hfill
 \begin{minipage}[b]{0.50\textwidth}
    \captionof{table}{Quantitative evaluation of volume fitting and compression.}
    \centering
        \begin{tabular}{@{}l@{\hskip 5pt}ccc}
        \toprule
           & Size (MB)  & SSIM  & PSNR \\ \midrule
        Raw, \texttt{float32}
        & 64 & - & -  \\
        \texttt{uint8} (zipped)   
        & 4.52 & 0.999 &  114.9    \\
        2D \texttt{jpeg} stack  
        & 0.92 & 0.981 & 44.5   \\
        FaCT-GS (Ours)
        &  0.84 & 0.976 & 44.2    \\ \bottomrule
        \end{tabular}
    \label{tab:compression}
    \end{minipage}
\end{minipage}
\end{figure}

Both the qualitative (\cref{fig:compression}) and quantitative (\cref{tab:compression}) results demonstrate that voxelizer-based fitting enables accurate representation of arbitrary volumes, with the Gaussian-based representation reaching a near-perfect average SSIM score of 0.976 -- much higher than any of the reconstruction methods from~\cref{tab:performance_basic}. 

The quantized Gaussian models require only 0.84 MB of storage, yielding a higher compression ratio than zipped \texttt{uint8} and a slightly higher ratio than a stacked \texttt{JPEG} representation. In terms of fidelity, the Gaussian representation exhibits a lower quality than \texttt{uint8} (which is effectively near-lossless), and performs comparably to the \texttt{JPEG} stack.

\section{Discussion and Limitations} \label{sec:dicsussion}

\textbf{Performance and scalability} \quad
The results in~\cref{sec:result_basic,sec:scaling_capability} demonstrate that FaCT-GS achieves superior reconstruction quality at a lower execution time, with the performance gap widening as projection and volume resolution increase. The ability to reconstruct high-resolution volumes within 1-4 minutes has a profound impact on the efficiency and scalability of CT imaging, enabling time-sensitive applications such as in-line industrial inspection or time-resolved acquisition.

While the observed gains stem primarily from the introduced optimizations, there is also a structural advantage inherent to GS-based reconstruction. Algebraic methods such as FISTA or SIRT iteratively update a discrete 3D voxel grid, making their computational cost dependent on both projection resolution and volume size. In contrast, GS-based reconstruction depends primarily on projection resolution -- through rasterization logic, while a complete voxelization may only be done once at the end of optimization. Although larger volumes may require more Gaussians, their number typically scales much more slowly than the total voxel count. A similar volume-independence applies to NeRFs. However, their significantly higher per-iteration cost makes this benefit less impactful.

An inherent limitation of the method is the requirement to manually select the number of Gaussians. Although adaptive control mitigates the potential negative impact of a sub-optimal choice, to our knowledge, no clear guidelines exist for determining the exact splat count for a given CT reconstruction task. At the same time, we observe that increasing the initial number of Gaussians often yields only marginal improvements. In contrast, algebraic methods do not require such design choices and may therefore remain attractive in certain scenarios, despite their lower reconstruction performance.

\textbf{Warm start impact} \quad
Our experiments in~\cref{sec:init_eval} focus on demonstrating that -- similarly to classical algebraic methods -- the proposed strategy initializes the optimization closer to a state with a high-quality solution. It is implicitly assumed that reaching that solution then requires fewer iterations than in the cold-start scenario. In practice, however, identifying non-trivial cases where this assumption can be confirmed is not always straightforward. Nevertheless, variants of warm-start strategies are successfully applied in many related contexts~\cite{chen2017a,NIU2018167,chen2008a}, and there exist several practical, albeit relatively trivial, scenarios where the lack of warm-start capability is a clear disadvantage. These include repetitive scans of identical or highly similar objects, long-term patient monitoring, or multi-resolution imaging. A small study on the impact of our warm start strategy on reconstruction speed is included in \emph{Supplementary Material}.

\textbf{Compact Gaussian representation} \quad
In~\cref{sec:fit_compress}, we demonstrate the volume fitting capabilities of FaCT-GS and show that the resulting Gaussian state has similar compression parameters to a 2D \texttt{JPEG} stack. This property is particularly relevant for CT data, where volumetric datasets frequently reach tens to hundreds of gigabytes. Moreover, prior work suggests that the Gaussian representation can be used directly for downstream processing and analysis~\cite{cen2025a,ye2025a,choi2025a}, further limiting the need for generating and storing the voxelized representation.

\section{Conclusion} \label{sec:conclusion}

Enabled by systematic optimization of the core rasterization, voxelization, and loss components, FaCT-GS consistently outperforms existing sparse-view reconstruction methods across both standard and high-resolution settings. This performance advantage, combined with the information-dense Gaussian representation for both reconstructed and pre-existing volumes, as well as the proposed warm-start capability, positions FaCT-GS as a practical, efficient, and versatile tool, ready to be adapted in modern CT workflows. Implementation and data used in the study can be found at \url{https://papieta.github.io/fact-gs/}.

\section{Acknowledgements}
The authors acknowledge funding from Villum Synergy through RENNER (REal-time acquisitioN and NEural Representation of structural properties), grant no. 00050097. Additional support was provided by the DEXTRA project (IFD3149-00001B) and the QUAITOM project (NNF21OC0069766). The authors also thank the 3D Imaging Center (3DIM) for its support in data acquisition. 


\bibliographystyle{splncs04}
\bibliography{main}

\input{supplementary}

\end{document}

%% file: preamble.tex
\usepackage{svg}
\usepackage{siunitx}
\usepackage[svgnames]{xcolor}
\usepackage[table]{xcolor}
\usepackage{arydshln}
\usepackage{makecell}

\newcommand{\nI}[1]{%
  \begingroup
  \setlength\fboxsep{1pt}%
  \colorbox{Salmon}{#1}%
  \endgroup
}
\newcommand{\nIc}{\cellcolor{Salmon}}

\newcommand{\nII}[1]{%
  \begingroup
  \setlength\fboxsep{1pt}%
  \colorbox{LightSalmon}{#1}%
  \endgroup
}
\newcommand{\nIIc}{\cellcolor{LightSalmon}}

\newcommand{\nIII}[1]{%
  \begingroup
  \setlength\fboxsep{1pt}%
  \colorbox{LightGoldenrodYellow}{#1}%
  \endgroup
}
\newcommand{\nIIIc}{\cellcolor{LightGoldenrodYellow}}

%% file: supplementary.tex
\renewcommand{\thesection}{\Alph{section}}

\clearpage

\title{Supplementary material for: FaCT-GS: Fast and Scalable CT Reconstruction with Gaussian Splatting} 

\titlerunning{FaCT-GS Supplementary Material}

\author{Pawel Tomasz Pieta \orcidlink{0009-0005-7634-6627} \and
Rasmus Juul Pedersen \orcidlink{0009-0007-8395-4270} \and
Sina Borgi \orcidlink{0000-0001-7404-7766} \and
Jakob Sauer Jørgensen \orcidlink{0000-0001-9114-754X} \and
Jens Wenzel Andreasen \orcidlink{0000-0002-3145-0229} \and
Vedrana Andersen Dahl \orcidlink{0000-0001-6734-5570} }

\authorrunning{P.T.~Pieta et al.}

\institute{Technical University of Denmark, 
Kgs. Lyngby, Denmark
\email{\{papi,rjupe,borgi,jakj,jewa,vand\}@dtu.dk}}

\maketitle

\input{lastcounters.tex}

\section{Performance ablation study}

In~\cref{sec:code_acceleration}, we introduce a collection of pipeline acceleration strategies, whose combined impact is summarized in~\cref{tab:performance_basic}. To better understand the contribution of each component, we perform an ablation study in which the proposed accelerations are individually disabled. The study is conducted in two scenarios: \textbf{[I]} -- the 75-view instance of the main experiment (\cref{tab:performance_basic}) on an RTX 4070 GPU, and \textbf{[II]} -- the 50-view 2k Coral experiment (\cref{fig:teaser_figure}) on an LS40 GPU. We first evaluate the major pipeline components, namely the fused TV and SSIM losses and the accelerated voxelizer and rasterizer modules. We then perform a fine-grained analysis of the rasterizer improvements, specifically the axis-aligned Gaussian bounding box (AA BBox) and the per-Gaussian backward pass.

The results (\cref{tab:GPU_component_ablation}) show that the largest speedup is achieved through rasterizer acceleration, driven primarily by the per-Gaussian backward pass. Although the remaining optimizations contribute less individually, they still provide substantial reductions in overall execution time. 

The trends are largely consistent across both scenarios, with most acceleration gains scaling predictably with problem size. The reduced impact of the voxelizer and TV acceleration can be attributed to both components operating on a fixed $32^3$ volume region. In this setting, the more powerful LS40 GPU registers a proportionally weaker acceleration. It is important to note, however, that when the voxelizer is applied in a volume fitting scenario, its acceleration impact is expected to be comparable to, or potentially greater than, that of the rasterizer during reconstruction.

\begin{table}[h!]
\caption{Impact of the most significant pipeline accelerations on execution time under two scenarios: \textbf{[I]} -- the 75-view instance of the main experiment (\cref{tab:performance_basic}) on an RTX 4070 GPU, and \textbf{[II]} -- the 50-view 2k Coral experiment (\cref{fig:teaser_figure}) on an LS40 GPU.}
\centering

\begin{tabular}{@{}c@{\hspace{16pt}}ccccc@{\hspace{16pt}}ccc@{}}
\toprule
 & \multicolumn{5}{c@{\hspace{20pt}}}{\makecell{\textbf{Reverted}\\\textbf{pipeline components}}}
 & \multicolumn{3}{c}{\makecell{\textbf{Reverted}\\\textbf{rasterizer components}}} \\
\cmidrule(l{-1pt}r{15pt}){2-6}\cmidrule(l{1pt}){7-9}
\makecell{Ablation\\scenario}
 & \makecell{None}
 & \makecell{Fused\\TV}
 & \makecell{Fused\\SSIM}
 & \makecell{Accel.\\Vox.}
 & \makecell{Accel.\\Rast.}
 & \makecell{AA\\BBox}
 & \makecell{Per-Gauss.\\backward}
 & \makecell{Others} \\
\midrule
\textbf{[I]}  & 161 s & +10 s & +34 s  & +54 s & +359 s  & +16 s  & +339 s  & +4 s  \\
\textbf{[II]} & 217 s & +6 s  & +187 s & +37 s & +3568 s & +155 s & +3556 s & +57 s \\
\bottomrule
\end{tabular}

\label{tab:GPU_component_ablation}
\end{table}

\section{Warm start impact}

In~\cref{sec:init_eval}, we evaluate the effectiveness of the proposed warm-start strategy and compare it with the approach used in established algebraic reconstruction methods. However, as highlighted in~\cref{sec:dicsussion}, we refrain from demonstrating the potential impact that an accurate volumetric prior may have on reconstruction speed. In this section, we supplement the previous analysis with a small study investigating reconstruction speed.

\begin{figure}[t]
\centering
\includegraphics[width=0.8\columnwidth]{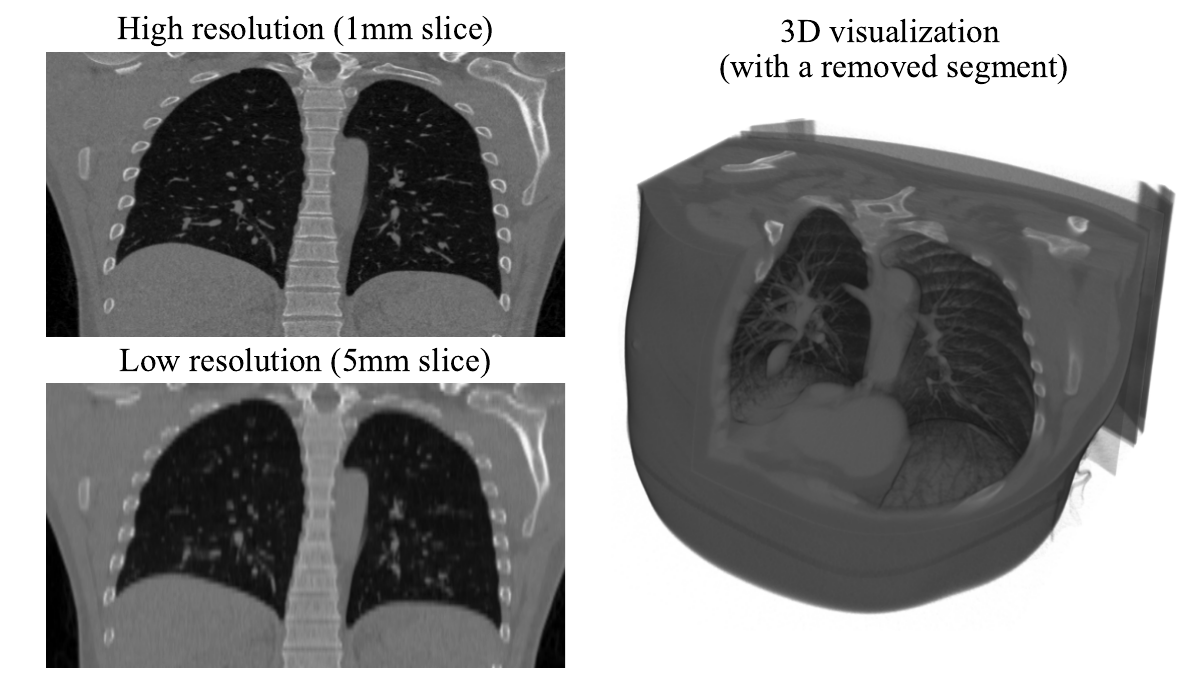}
\caption{Example scan pair used in the hyperresolution warm-start study, and a 3D volume visualization. Two scans of the same patient form a pair, where the high-resolution scan serves as the reconstruction target and the low-resolution scan provides the volumetric prior.}
\label{fig:hyperres_data}
\end{figure}

We use the RPLHR-CT hyperresolution dataset~\cite{yu2022a} containing pairs of medical chest CT scans. Each pair consists of two scans of the same patient acquired at different longitudinal resolutions: 1 mm for the high-resolution instance and 5 mm for the low-resolution instance (example shown in~\cref{fig:hyperres_data}). The sagittal and frontal dimensions remain unchanged at a resolution of $512^2$. The number of longitudinal slices varies slightly across scans, typically staying around 280 for the high-resolution volumes.

Five scan pairs are selected from the dataset. Each volume is resized to a common size of $280\times512\times512$ and padded to form a $512^3$ cube. From the high-resolution volumes, we generate 50 synthetic projections of size $1024^2$, which serve as reconstruction targets. The corresponding low-resolution scans are used as volumetric priors. Similarly to~\cref{sec:init_eval}, the volume fitting pipeline is executed for 50 iterations. However, the number of Gaussians is increased to 100{,}000 to account for the larger size of the target volume.

For each pair, a full reconstruction ($N_\mathrm{iter} = 400$) is performed in both cold-start and warm-start scenarios, and the 3D SSIM and PSNR scores obtained at the end of each reconstruction are recorded. Finally, we also report the normalized iteration ratio $\tau_\mathrm{iter}$ [\%], defined as the fraction of total iterations required in the warm-start scenario to reach the maximum metric score (PSNR or SSIM) achieved in the cold-start reconstruction:
\begin{equation}
    \tau_\mathrm{iter} = \frac{n_\mathrm{match, metric_{max}}}{N_\mathrm{iter}}.
\end{equation}

\begin{table*}[t!]
\caption{Quantitative impact of volume prior warm start on the reconstruction quality. Warm-started reconstructions achieve slightly higher scores and reach comparable performance in significantly fewer iterations.}
\centering
\begin{tabular}{@{}c@{\hspace{10pt}}cc@{\hspace{10pt}}cc@{\hspace{10pt}}cc}
\toprule
    & \multicolumn{2}{c@{\hskip 10pt}}{Cold start}  & \multicolumn{2}{c@{\hskip 10pt}}{Warm start}  & \multicolumn{2}{c}{$\tau_\mathrm{iter}$ [\%]} \\  \cmidrule(l{-1pt}r{10pt}){2-3}\cmidrule(l{-1pt}r{10pt}){4-5}\cmidrule(l{-1pt}r{-0.5pt}){6-7}
Index & PSNR & SSIM  & PSNR & SSIM & PSNR & SSIM  \\ \midrule 

1
& 34.74 &  0.924 &  35.03  & 0.927 & 54.0 &  58.0  \\ 
2
& 34.94 &  0.916 &  35.19  & 0.920 & 63.0 &  56.0  \\
3
& 34.10 &  0.908 &  34.36 & 0.912 & 59.5 &  47.5  \\
4
& 34.99 &  0.921 &  35.18  & 0.922 & 68.0 &  77.0  \\
5
& 33.80 &  0.902 &  33.96  & 0.904 & 69.0 &  67.5  \\ \midrule
Avg 
& 34.51 & 0.914 & 34.74 & 0.917 & 62.7 & 61.2 \\ \bottomrule
\end{tabular}
\label{tab:hyperres_result}
\end{table*}

The results of the hyperresolution warm-start study are reported in~\cref{tab:hyperres_result}. In all cases, the warm-started reconstruction achieves slightly higher PSNR and SSIM scores. More importantly, the average $\tau_\mathrm{iter}$ is 62.7\% for PSNR and 61.2\% for SSIM. This indicates that the prior-based reconstruction requires nearly 40\% fewer iterations to reach the quality achieved in the cold-start scenario.

\section{Gaussian number ablation study}
As discussed in~\cref{sec:dicsussion}, the performance of GS-based reconstruction methods depends on the number of Gaussians used during optimization, which can complicate practical use. To develop an intuitive understanding of which image properties impact the required number of Gaussians, we conduct an ablation study using all volumes from the R$^2$-Gaussian dataset as well as the multi‑resolution coral volumes used in the scaling experiment in~\cref{sec:scaling_capability}. For each volume, we evaluate how the Gaussian count affects reconstruction quality. We run 400 reconstruction iterations on 50 input views, while varying both the initial and maximum number of Gaussians, with the maximum set to 10\% above the initial value.

To assess when a reconstruction problem reaches a sufficient Gaussian saturation, we compute a relative SSIM score defined as:
\begin{equation}
    \mathrm{SSIM}_\mathrm{rel} = \frac{\mathrm{SSIM}}{\mathrm{SSIM_{MAX}}}, 
\end{equation}
where $\mathrm{SSIM_{MAX}}$ is the highest SSIM obtained for each volume across all tested initial Gaussian counts. We use a 99\% threshold as a practical criterion indicating that any further improvements are negligible.

\begin{figure*}[!t]
\centering
     \begin{subfigure}[b]{0.49\columnwidth}
         \includegraphics[width=\columnwidth]{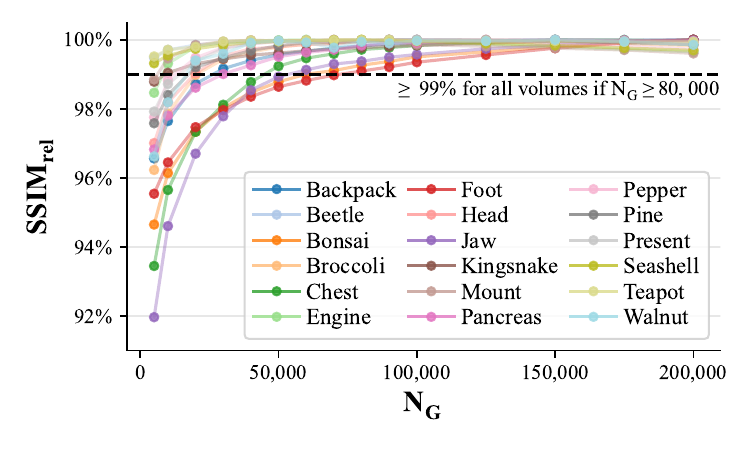}
        \caption{Varying volume type. R$^2$-Gaussian dataset~\cite{zha2024r_suppl}.}
        \label{fig:r2_data_gauss_num}
     \end{subfigure}
     \begin{subfigure}[b]{0.49\columnwidth}
         \includegraphics[width=\columnwidth]{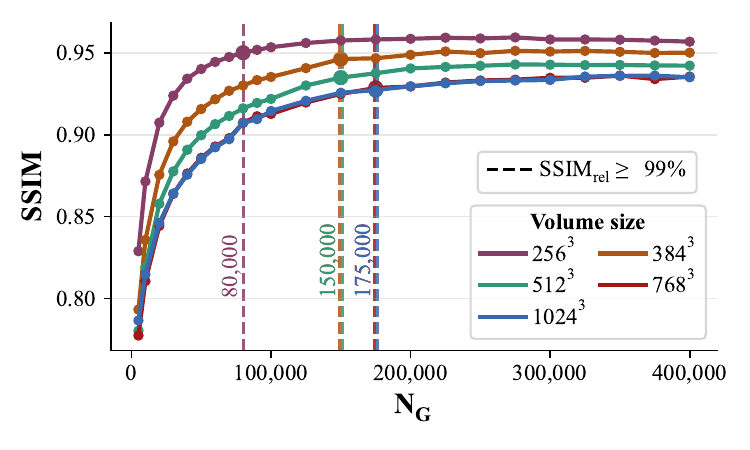}
        \caption{Varying volume size. Multi-resolution coral data.}
        \label{fig:coral_gauss_num}
     \end{subfigure}
\caption{Gaussian number ablation study for varying type and resolution of reconstructed data. Complex, noisy, or large volumes can require significantly more Gaussians to reach maximum possible reconstruction quality. Increasing Gaussians beyond a certain point tends to yield diminishing returns.}
\label{fig:gauss_num_vols}
\end{figure*}

In~\cref{sec:dicsussion} we highlight that in our experience, increasing the Gaussian count generally tends to produce diminishing returns. \cref{fig:r2_data_gauss_num} confirms this trend -- all but three volumes reach the 99\% threshold with 50,000 initial Gaussians, and all volumes surpass it by 80,000. Most volumes, however, reach the threshold well before these values.

\begin{figure*}[!t]
\centering
     \begin{subfigure}[b]{0.46\columnwidth}
         \includegraphics[width=\columnwidth]{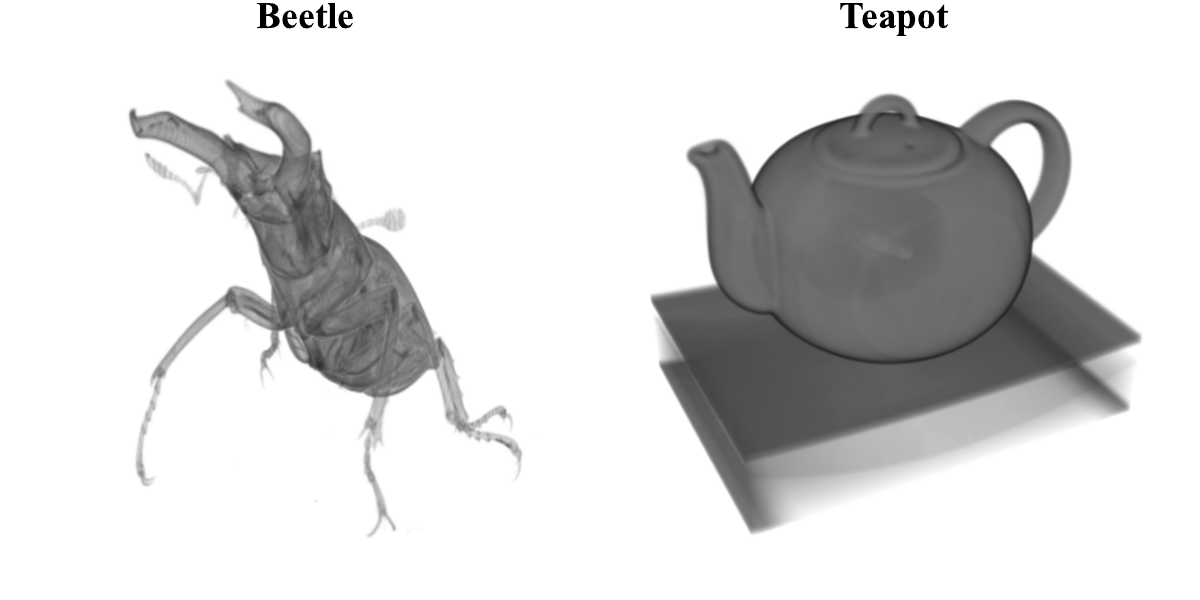}
        \caption{Beetle and Teapot required only 5,000 Gaussians.}
        \label{fig:gauss_num_vols_good}
     \end{subfigure}
     \hspace{0.5cm}
     \begin{subfigure}[b]{0.46\columnwidth}
         \includegraphics[width=\columnwidth]{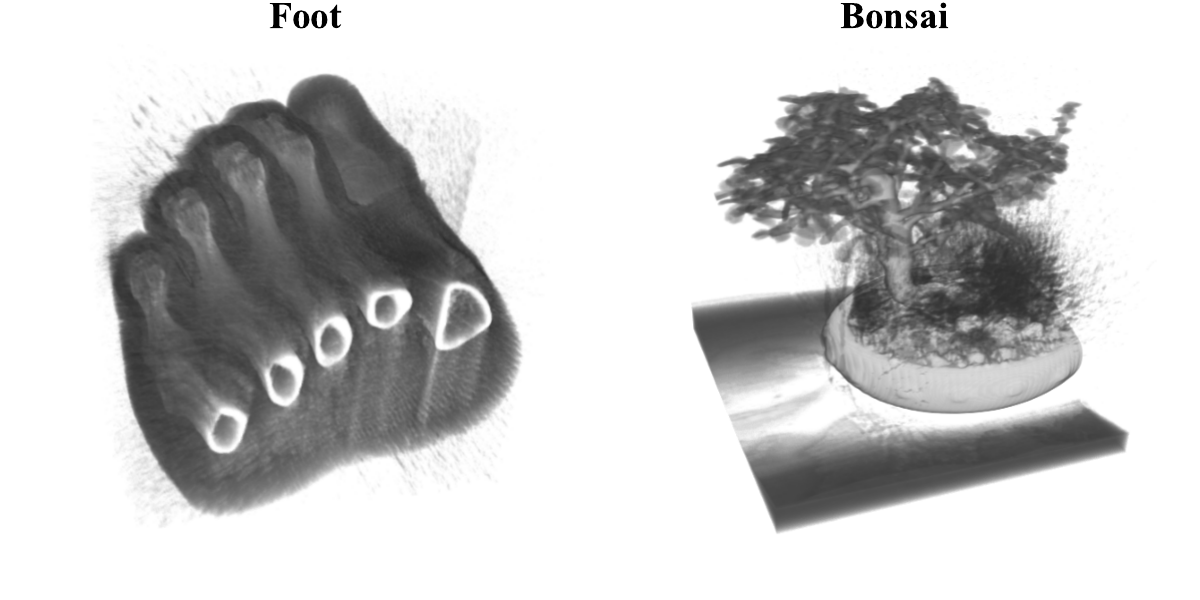}
        \caption{Foot and Bonsai required at least 80,000 Gaussians.}
        \label{fig:gauss_num_vols_bad}
     \end{subfigure}
\caption{Example volumes from the R$^2$-Gaussian dataset that required the least and most Gaussians to reach the 99\% reconstruction quality threshold. Noise, structural complexity, and varying density are the primary factors leading to higher Gaussian count requirements (assuming the same volume size).}
\label{fig:gauss_num}
\end{figure*}

\cref{fig:gauss_num} further illustrates differences in Gaussian requirements by showing examples of volumes at opposite ends of this spectrum. The Teapot and Beetle volumes reach a high and stable $\mathrm{SSIM}_\mathrm{rel}$ with just 5,000 Gaussians, reflecting their predominantly smooth surfaces and limited structural complexity. In contrast, the Bonsai and Foot volumes require at least 80,000 Gaussians to reach the 99\% threshold, likely due to the increased number of fine‑scale features. The Bonsai scan contains irregular, complex details, while the Foot volume suffers from noise commonly observed in mixed hard‑ and soft‑tissue scans.

In~\cref{sec:scaling_capability}, we argue that an increase in volume size leads to a proportional increase in the represented level of detail, suggesting the need for a larger number of Gaussians. The second part of the study -- based on the multi-resolution coral data -- provides further insight into this hypothesis. Results in~\cref{fig:coral_gauss_num} show that as the resolution increases, the 99\% threshold shifts to the right, confirming that larger volumes require more Gaussians to reach the practical plateau. Notably, the $768^3$ and $1024^3$ volumes achieve almost identical SSIM scores at each initial Gaussian count, suggesting that size alone is not a sufficient indicator of detail or structural complexity. We hypothesize that beyond a certain resolution, all details are sufficiently represented, and further increases in volume size merely enlarge those structures, requiring larger Gaussians rather than a greater number of them.

The results in \cref{fig:coral_gauss_num} also reveal an important nuance in reconstructing high-resolution volumes. With the increase in resolution, the maximum achievable SSIM slightly drops. If the goal is only to reach the 99\% convergence threshold, the Gaussian count does not need to scale strongly with resolution, and relatively modest increases are often sufficient. However, if the goal is to maintain or approach the absolute SSIM obtained at lower resolutions, then a substantial increase in Gaussian count does help to narrow this gap.

\section{Qualitative reconstruction results}

\begin{figure}[h]
\centering
\includegraphics[width=0.98\columnwidth]{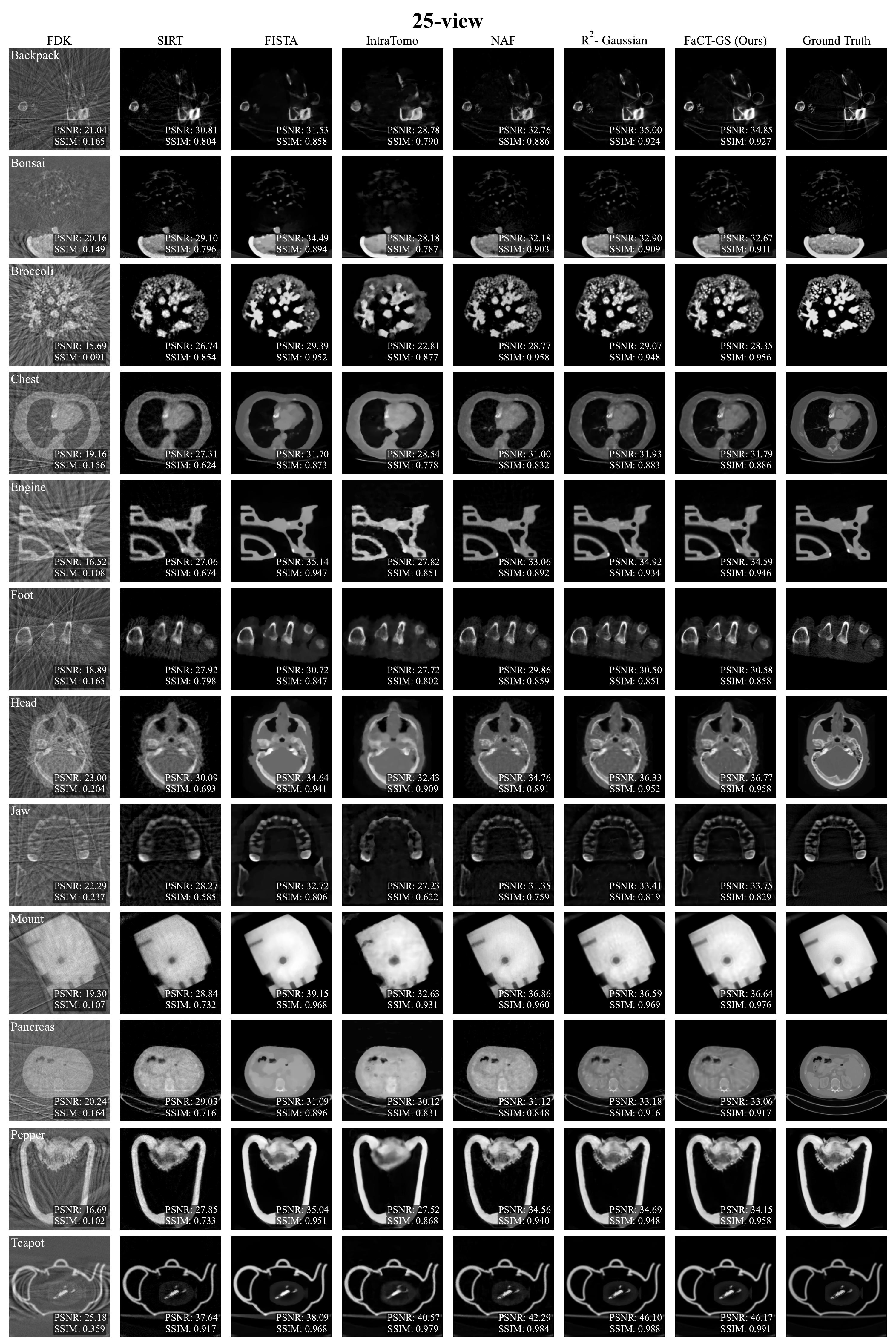}
\caption{Qualitative comparison of reconstruction quality on select volumes from the 25-view setting of the R$^2$-Gaussian\cite{zha2024r_suppl} dataset.}
\label{fig:recon_comp_25}
\end{figure}

\begin{figure}[h]
\centering
\includegraphics[width=0.98\columnwidth]{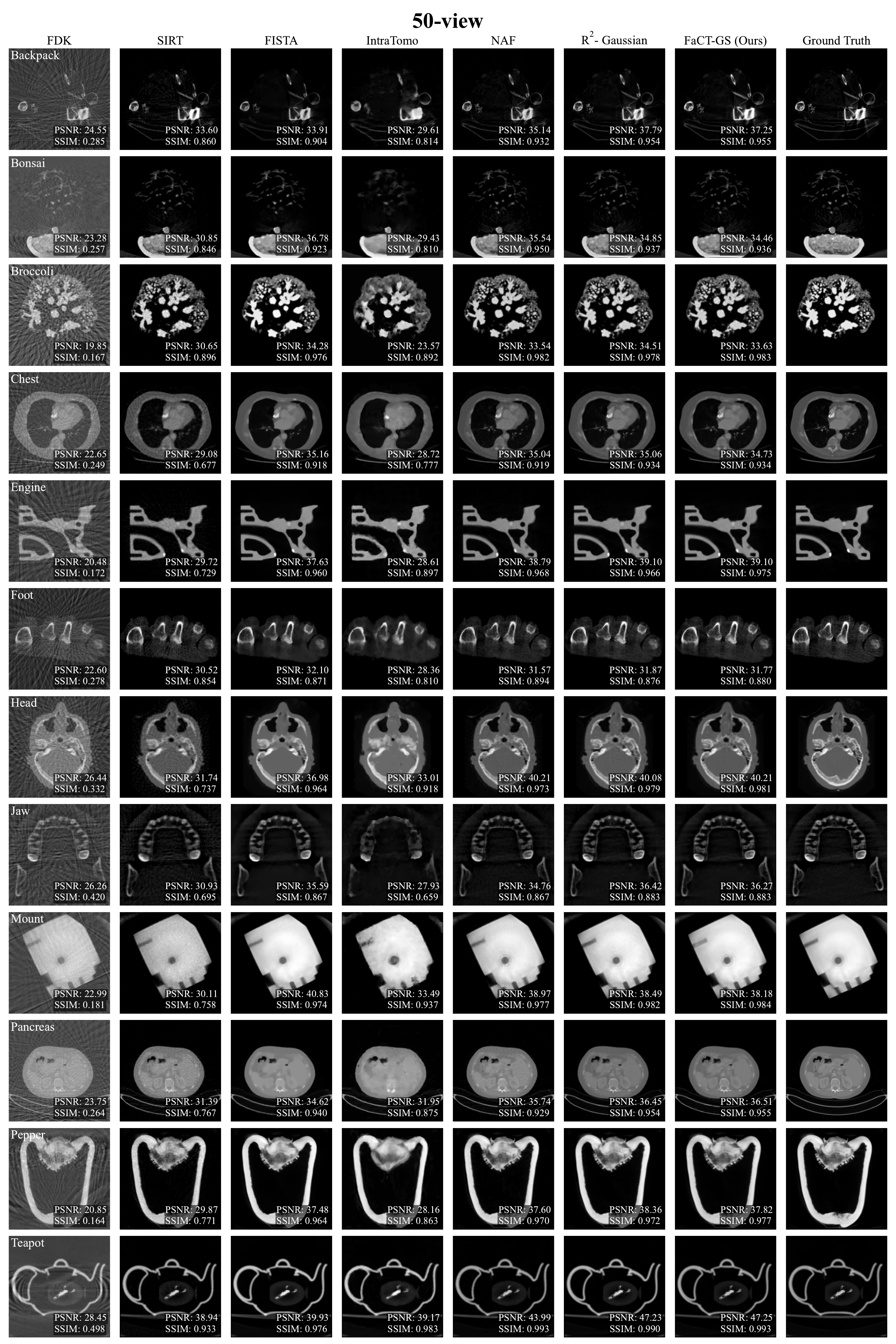}
\caption{Qualitative comparison of reconstruction quality on select volumes from the 50-view setting of the R$^2$-Gaussian\cite{zha2024r_suppl} dataset.}
\label{fig:recon_comp_50}
\end{figure}

\begin{figure}[h]
\centering
\includegraphics[width=0.98\columnwidth]{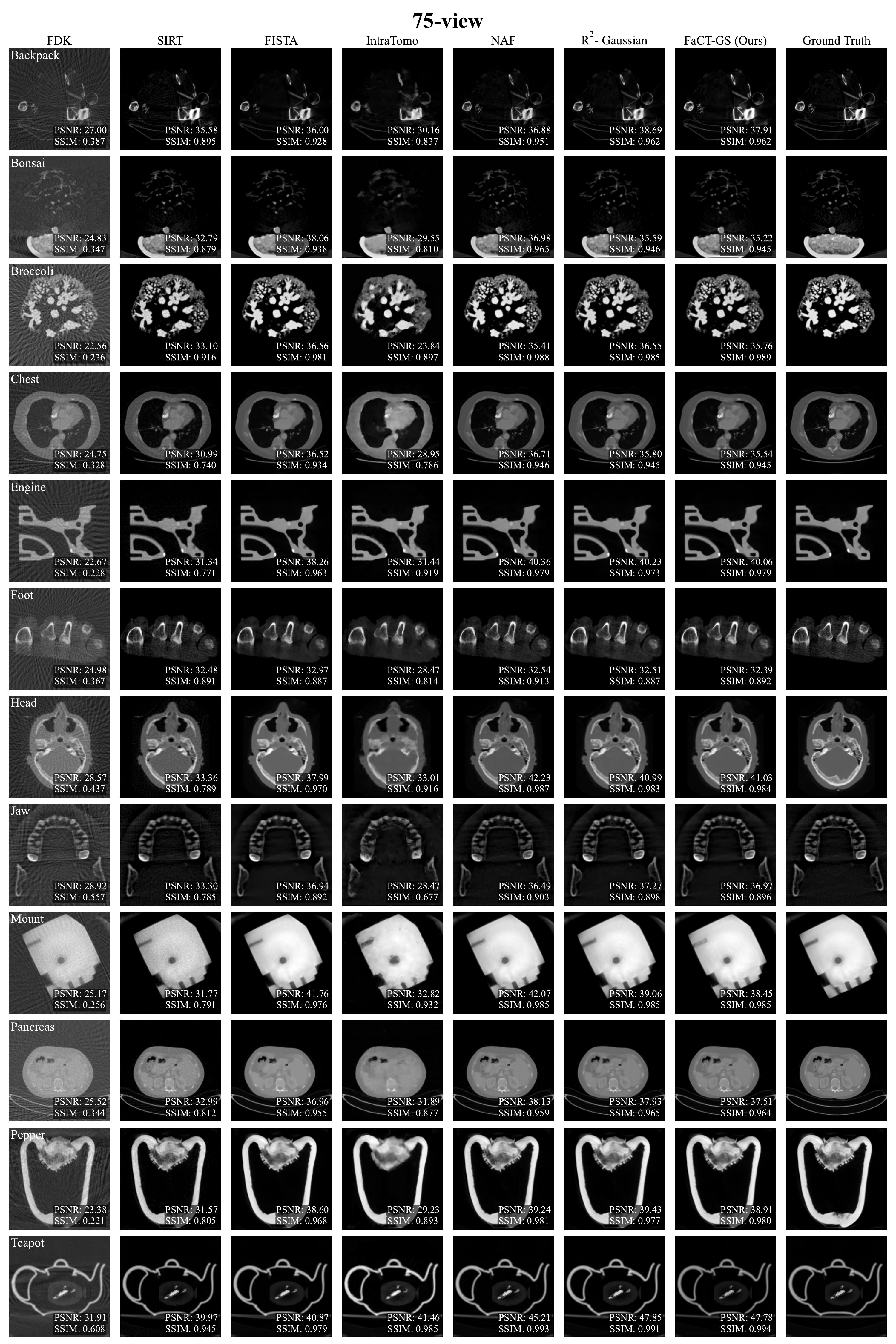}
\caption{Qualitative comparison of reconstruction quality on select volumes from the 75-view setting of the R$^2$-Gaussian\cite{zha2024r_suppl} dataset.}
\label{fig:recon_comp_75}
\end{figure}

In~\cref{sec:result_basic} and~\cref{tab:performance_basic}, we present a quantitative reconstruction evaluation on the R$^2$-Gaussian dataset~\cite{zha2024r_suppl}. To complement these results, we provide a qualitative comparison of slices from selected volumes reconstructed with all tested methods, shown in Figs.~\ref{fig:recon_comp_25}, \ref{fig:recon_comp_50}, and \ref{fig:recon_comp_75} for the 25-, 50-, and 75-view settings, respectively. 

Qualitative results confirm the trends reported in \cref{tab:performance_basic}. Both R$^2$-Gaussian and FaCT-GS outperform their predecessors in terms of SSIM and PSNR, with FISTA and NAF being the closest competitors. Visually, reconstructions produced by the GS-based methods are smoother than those from NAF and SIRT, but sharper than those from FISTA, resulting in a level of detail that most closely matches the ground-truth volume. However, GS-based methods appear to be more sensitive to noisy data (as seen, e.g., in the Foot scan), tending to overfit to anisotropic streak artifacts more than the other methods.

%% file: lastcounters.tex
\setcounter {figure}{6}
\setcounter {table}{2}
\setcounter {equation}{6}